\documentclass[sigconf]{acmart}

\usepackage{amsmath}
\usepackage{booktabs}
\usepackage{bm}
\usepackage{multicol}
\usepackage{multirow}
\usepackage{amssymb}

\usepackage{mathtools}
\usepackage{amsthm}

\usepackage{xspace}
\usepackage{fontawesome}
\usepackage{mathrsfs}

\usepackage{colortbl}

\usepackage{subfig}
\usepackage{graphicx}

\usepackage{tabularx}
\usepackage{arydshln}
\usepackage{makecell}
\usepackage{CJKutf8}

\usepackage{array}

\usepackage{wrapfig}

\usepackage{subcaption}
\usepackage{diagbox}

\definecolor{deepred}{rgb}{0.698,0.133,0.133}
\definecolor{blue}{rgb}{0,0,1}

\definecolor{myred}{rgb}{1.0, .349, .3686}
\definecolor{CoINBlue}{rgb}{0.129, 0.6196, 0.7373}     
\definecolor{CoINRed}{rgb}{0.8392, 0.1569, 0.1569}   
\definecolor{shadecolor}{rgb}{0.92,0.92,0.92}

\usepackage{caption}       

\usepackage[dvipsnames]{xcolor}

\usepackage[capitalize,noabbrev]{cleveref}

\usepackage{algorithm}

\usepackage{algpseudocode}

\AtBeginDocument{%
  \providecommand\BibTeX{{%
    \normalfont B\kern-0.5em{\scshape i\kern-0.25em b}\kern-0.8em\TeX}}}

\copyrightyear{2026}
\acmYear{2026}
\setcopyright{cc}
\setcctype{by}
\acmConference[MM '26]{Proceedings of the 34th ACM International Conference on Multimedia}{November 10--14, 2026}{Rio de Janeiro, Brazil}
\acmBooktitle{Proceedings of the 34th ACM International Conference on Multimedia (MM '26), November 10--14, 2026, Rio de Janeiro, Brazil}
\acmDOI{10.1145/3767308.3836259}
\acmISBN{979-8-4007-2213-4/2026/11}

\acmSubmissionID{7536}

\begin{document}

\title{Progressive Multimodal Alignment for Continual Instruction~Tuning}

\author{Duzhen Zhang}
\authornote{Both authors contributed equally to this research.}
\orcid{0000-0002-4280-431X}
\affiliation{%
  \institution{Mohamed bin Zayed University of Artificial Intelligence}
  \city{Abu Dhabi}
  \country{United Arab Emirates}
}
\affiliation{%
  \institution{Center for Excellence in Brain Science and Intelligence Technology, Chinese Academy of Sciences}
  \city{Shanghai}
  \country{China}
}
\email{duzhen.zhang@mbzuai.ac.ae}

\author{Yahan Yu}
\orcid{0000-0003-1610-1167}
\authornotemark[1]
\affiliation{%
  \institution{Kyoto University}
  \city{Kyoto}
  \country{Japan}
}
\email{yahan@nlp.ist.i.kyoto-u.ac.jp}

\author{Qiaoyi Su}
\orcid{0009-0002-4442-7391}
\affiliation{%
  \institution{Migu Culture Technology Co.,Ltd.}
  \city{Beijing}
  \country{China}
}
\email{suqiaoyi@migu.chinamobile.com}

\author{Jiahua Dong}
\orcid{0000-0001-8545-4447}
\affiliation{%
  \institution{Mohamed bin Zayed University of Artificial Intelligence}
  \city{Abu Dhabi}
  \country{United Arab Emirates}
}
\email{dongjiahua1995@gmail.com}

\author{Tielin Zhang}
\authornote{Corresponding author.}
\orcid{0000-0002-5111-9891}
\affiliation{%
  \institution{Center for Excellence in Brain Science and Intelligence Technology, Chinese Academy of Sciences}
  \city{Shanghai}
  \country{China}
}
\affiliation{%
  \institution{State Key Laboratory of Brain Cognition and Brain-inspired Intelligence Technology
}
  \city{Shanghai}
  \country{China}
}
\email{zhangtielin@ion.ac.cn}

\renewcommand{\shortauthors}{Duzhen Zhang, Yahan Yu, Qiaoyi Su, Jiahua Dong, and Tielin Zhang}

\begin{abstract}
Multimodal Large Language Models (MLLMs) rely on a projector to align visual representations with the language embedding space, making it central to cross-modal understanding. 
In Multimodal Continual Instruction Tuning (MCIT), however, shifting visual distributions and evolving instruction semantics cause this shared projector to drift, leading to projector-level forgetting, an issue largely overlooked by methods that focus primarily on the LLM backbone. 
We introduce Progressive Multimodal Alignment (PMA), a framework that enables the projector to adapt continually while preserving previously learned alignment. 
PMA detects multimodal distribution shifts via a lightweight representation descriptor and progressively expands projector experts only when needed. 
An expandable router integrates expert outputs based on multimodal features, while the original pretrained projector is retained as a stable alignment anchor. 
This progressive mechanism balances stability and plasticity with sub-linear parameter growth and serves as a method-agnostic add-on to existing MCIT approaches. 
Extensive experiments on two recent MCIT benchmarks demonstrate that mitigating projector-level forgetting yields consistent gains over prior state-of-the-art methods when combined with PMA. 
Moreover, PMA scales across diverse MLLM backbones, demonstrating robust and broadly applicable MCIT performance.\footnote{The code is available at \url{https://github.com/BladeDancer957/PMA}.}
\end{abstract}



\maketitle

\section{Introduction}

Multimodal Large Language Models (MLLMs) have advanced visual-language understanding and instruction following by coupling strong visual encoders with LLMs capable of open-ended reasoning \cite{DBLP:conf/icml/0008LSH23,yin2024survey,wu2023multimodal}.
A central component in this architecture is the projector, which maps visual features into the language embedding space and serves as the semantic interface for cross-modal alignment.

\begin{figure}[t!]
\centering
  \includegraphics[width=1.0\linewidth]{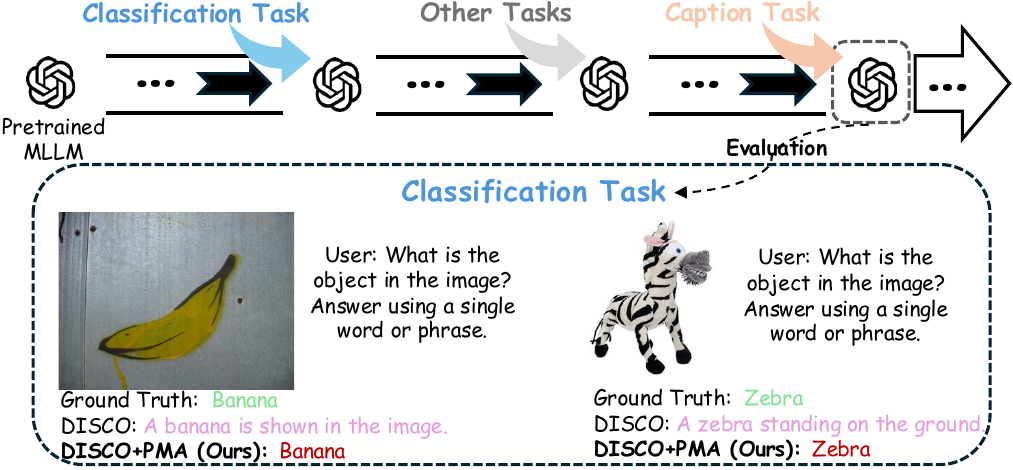}
  \caption{Illustration of projector-level forgetting in MCIT.
After finetuning on a captioning task, DISCO \cite{guo2025federated} produces caption-style responses even for classification instructions, indicating degraded visual translation for earlier tasks.
In contrast, DISCO+PMA (Ours) preserves task-specific cross-modal alignment and generates task-consistent classification outputs for the same inputs.}
\label{fig:intro}
\end{figure}

As MLLMs are increasingly deployed in dynamic environments, they must adapt to evolving tasks and instruction styles rather than operate as static, once-trained models. 
However, retraining them for every new task is prohibitively costly, motivating growing interest in Multimodal Continual Instruction Tuning (MCIT) \cite{he2023continual,chen2024coin,cao2024continual}, where models aim to acquire new capabilities while preserving both previously learned skills and established cross-modal alignment. 
Achieving this balance is challenging due to Catastrophic Forgetting (CF) \cite{mccloskey1989catastrophic,kirkpatrick2017overcoming,goodfellow2013empirical,dong2023federated,10323204}, as shifts in visual distributions and instruction semantics can cause models to lose prior knowledge and misalign earlier visual-language mappings.

To mitigate CF, existing MCIT methods primarily incorporate Parameter-Efficient FineTuning (PEFT) techniques, such as prompt tuning \cite{lester2021power,zeng2024modalprompt} and LoRA \cite{hu2022lora,yu2025progressive}, into the LLM backbone to preserve previously acquired capabilities.
For instance, MoELoRA \cite{chen2024coin} employs multi-expert LoRA architectures \cite{dou2023loramoe,liu2023moelora,zhang2025enhancing} to capture task-specific knowledge; 
HiDE \cite{DBLP:conf/acl/GuoZXZWZL25} leverages layer-wise similarity variations to decompose LoRA into task-specific expansion and task-general fusion components, striking a balance between adaptation performance and memory efficiency; 
and DISCO \cite{guo2025federated} assigns task-specific LoRA subspaces with subspace-selective activation to reduce interference, achieving State-Of-The-Art (SOTA) MCIT performance.
However, these approaches largely overlook the preservation of cross-modal alignment learned in earlier tasks, implicitly treating the projector as a shared, jointly finetuned module across all tasks. 
As a result, updating this shared projector induces alignment drift, whereby newly learned tasks overwrite previously established vision-language mappings, a phenomenon we term \textit{projector-level forgetting}.

Taking DISCO \cite{guo2025federated} as an illustrative example, after finetuning on an image captioning task, the model may exhibit a bias toward the most recent instruction style, causing incorrect translation of visual features for earlier image classification tasks.
As illustrated in Figure \ref{fig:intro}, we observe cases where the model produces caption-style responses even when instructed to output a concise classification label.
This behavior reflects an instruction-driven collapse in visual translation, where the projector fails to condition its mappings on earlier task instructions, leading to degraded cross-modal alignment and projector-level forgetting.
In contrast, when equipped with our method, the same inputs are routed to appropriate projector experts, and the model generates task-consistent classification outputs, avoiding this failure mode.

A naive solution is to allocate a separate projector for each task. 
However, this strategy is parameter-inefficient, scales linearly with the number of tasks, hinders knowledge sharing among related tasks, and requires task IDs at inference, which is incompatible with realistic MCIT settings where task IDs are unavailable. 
This leads to a fundamental question: \textit{How can we enable the projector to preserve its original alignment while adapting to new multimodal tasks in a parameter-efficient, transferable, and task-ID-free manner?}

In this paper, we address this challenge with Progressive Multimodal Alignment (PMA), a framework that allows the projector to adapt continually while retaining previously learned alignment. 
PMA detects multimodal distribution shifts via a lightweight Representation Descriptor (RD) and expands new projector experts only when necessary, promoting knowledge sharing and ensuring sub-linear parameter growth. 
An expandable router dynamically integrates expert outputs based on the same multimodal features as RD, enabling automatic routing without task-ID supervision, while the frozen pretrained projector provides a stable alignment anchor. 
Importantly, PMA is method-agnostic and integrates seamlessly with existing MCIT approaches, complementing their focus on the LLM backbone by directly addressing the long-overlooked issue of projector-level forgetting.

Our contributions can be summarized as follows:
\begin{itemize}
    \item We identify and formalize \textit{projector-level forgetting} as a key yet largely overlooked bottleneck in MCIT, demonstrating that drift in the projector undermines instruction retention.

   \item  We propose PMA, a method-agnostic framework that enables the projector to adapt continually by detecting multimodal distribution shifts, selectively expanding lightweight experts, and dynamically routing them without task IDs, all while preserving previously learned alignment with sub-linear parameter growth. 

    \item  We conduct extensive experiments on two MCIT benchmarks (UCIT \cite{DBLP:conf/acl/GuoZXZWZL25} and MLLM-DCL \cite{zhao2025mllm}), showing that PMA consistently improves previous SOTA methods and scales effectively across diverse MLLM backbones (LLaVA-1.5 \cite{liu2023improved}, InternVL \cite{chen2023internvl}).

\end{itemize}

\section{Related Work} \label{sec:rw}

\subsection{MLLMs}

Recent progress in MLLMs has significantly advanced visual-language understanding \cite{chen2023vlp} and instruction following \cite{zhang2024mm}. 
Early models such as BLIP-2 \cite{DBLP:conf/icml/0008LSH23} employ a frozen LLM paired with a frozen visual encoder and a learnable projector (\emph{e.g.}, Q-Former) to achieve efficient modality alignment. 
Subsequent systems, including LLaVA \cite{liu2023llava}, MiniGPT-4 \cite{zhu2023minigpt}, and QwenVL \cite{bai2023qwen}, simplify the alignment mechanism using linear projectors and show that instruction tuning plays a crucial role in aligning visual features with human intent. 
Recent variants, such as LLaVA-1.5 \cite{liu2023improved}, ShareGPT4V \cite{chen2023sharegpt4v}, and InternVL \cite{chen2023internvl}, further refine these alignment strategies and show strong performance across a diverse set of multimodal benchmarks. 
Meanwhile, the MLLM ecosystem has expanded beyond static images to modalities such as video and audio \cite{gpt4o-0513,lillava,liu2024llavanext,bai2025qwen2,fu2025vita}, signaling a broader shift toward more general-purpose multimodal reasoning. 
However, as model scale and application complexity continue to grow, adapting MLLMs to evolving tasks and instruction styles without retraining from scratch becomes both necessary and challenging. 
This demands new paradigms for MCIT, enabling MLLMs to maintain alignment with human intent in dynamic, real-world environments.

\subsection{MCIT} 

Building on the need for adaptable MLLMs, recent work has begun to explore MCIT, which aims to maintain alignment as instruction styles and task distributions evolve. 
To support systematic evaluation, several benchmarks have been proposed \cite{cao2024continual,he2023continual}. 
CoIN \cite{chen2024coin} and UCIT \cite{DBLP:conf/acl/GuoZXZWZL25} introduce dataset-incremental settings; however, CoIN suffers from pretraining overlap that leads to information leakage, while UCIT addresses this by selecting datasets minimally correlated with LLaVA's pretraining data. 
More recent efforts like MLLM-DCL \cite{zhao2025mllm} further broaden the benchmark landscape with domain-specific knowledge.

\begin{figure*}[t!]
\centering
  \includegraphics[width=1.0\linewidth]{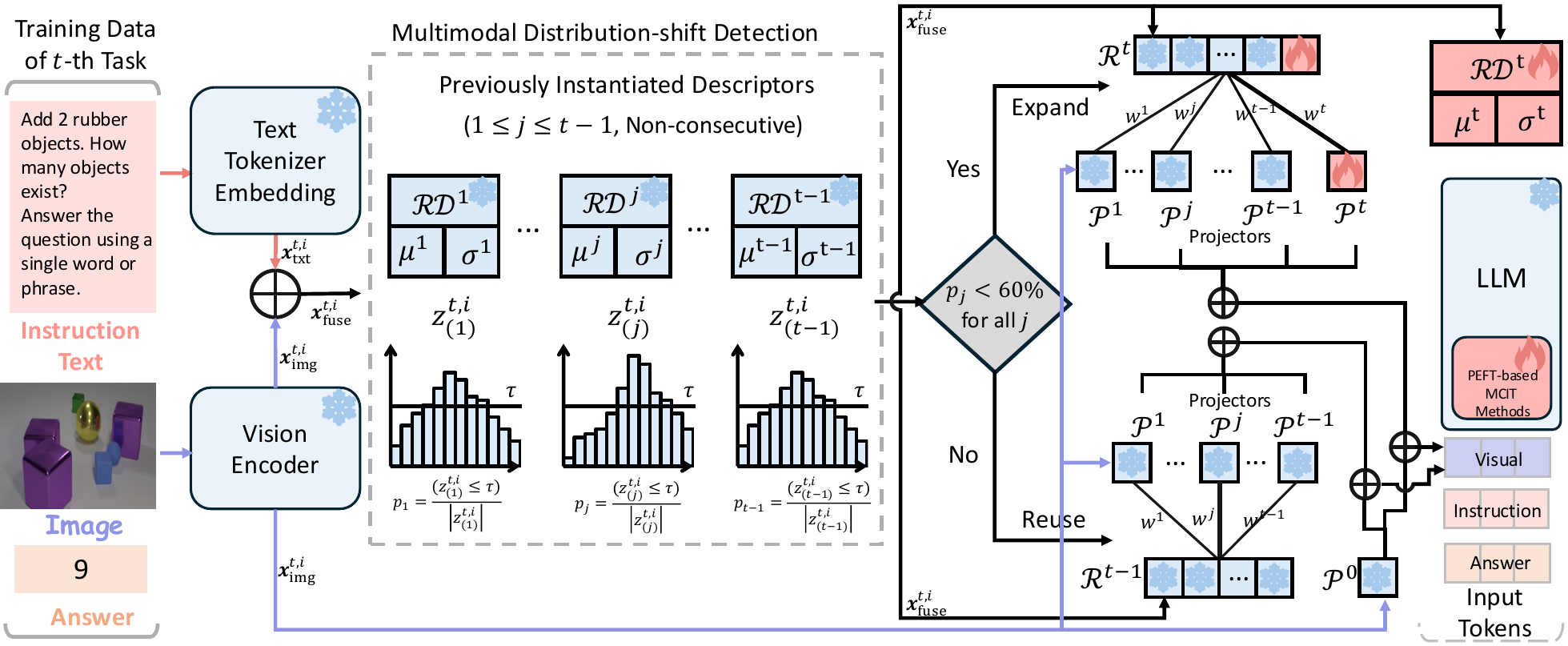}
    \caption{Overview of PMA. 
    Lightweight RDs detect multimodal distribution shifts and determine whether to trigger projector expansion or reuse existing projector experts. 
    An expandable router ($\mathcal{R}^t$) mixes projector experts without task IDs, while the frozen pre-trained projector ($\mathcal{P}^0$) serves as a stable alignment anchor. 
    PMA is method-agnostic and integrates seamlessly with existing PEFT-based MCIT methods that primarily focus on the LLM side, enabling progressive projector-side adaptation with sub-linear parameter growth.}
\label{fig:method}
\end{figure*}

To mitigate CF, recent research work adapts various PEFT strategies to the MCIT setting \cite{gedynamic,zhang2023brain,zheng2026lifelong,zheng2025lifelongagentbench}. 
MCITlib \cite{guo2025mcitlib} provides a unified framework that consolidates representative approaches such as MoELoRA \cite{chen2024coin} maintains multiple LoRA experts to capture task-specific information; 
SEFE \cite{chensefe} addresses both superficial and essential forgetting by harmonizing task styles via answer style diversification and stabilizing critical parameters with RegLoRA; 
and DISCO \cite{guo2025federated} allocates task-specific LoRA subspaces during training and employs subspace-selective activation during inference to reduce interference. 
More method introductions are provided in Section \ref{exp:baselines}.

While existing MCIT methods primarily address CF in the LLM backbone, projector-level forgetting remains largely overlooked. 
To address this gap, we propose PMA, which expands representational capacity only when needed while preserving previously learned alignment. 
By directly targeting projector-level forgetting, an underexplored but critical bottleneck in MCIT, PMA integrates seamlessly with nearly all existing PEFT-based MCIT methods.

\section{Method} \label{sec:method}

We propose PMA, a method-agnostic framework that complements existing MCIT approaches by explicitly addressing projector-level forgetting, an issue largely overlooked by methods that focus on the LLM backbone.
As illustrated in Figure \ref{fig:method}, PMA employs lightweight RDs to detect multimodal distribution shifts and expands the projector with a new expert only when such shifts are detected.
An expandable router integrates expert outputs based on the same multimodal features used by the RD, enabling task-agnostic inference and facilitating knowledge sharing across related tasks, while the frozen pretrained projector is retained as a stable alignment anchor.
Overall, PMA provides a progressive and parameter-efficient mechanism for maintaining cross-modal alignment, and can be seamlessly integrated with prior PEFT-based MCIT methods.

\subsection{Task Formulation}

MCIT \cite{guo2025mcitlib} aims to update an MLLM with new instruction-driven tasks without incurring the cost of full retraining. 
We consider a setting where an MLLM is finetuned over a sequence of tasks $t = {1, \dots, T}$, each associated with a training set $\mathcal{D}^{t}_\text{train}$ and a test set $\mathcal{D}^{t}_\text{test}$. 
Each instance $x^{t,i}$ in these datasets consists of an image $x^{t,i}_{\text{img}}$, a text instruction $x^{t,i}_{\text{txt}}$, and an answer $x^{t,i}_{\text{ans}}$. 
The goal is to incrementally adapt a single model $\mathcal{M}$ while preserving strong performance on all previously learned tasks. 
MCIT is typically evaluated in a \textbf{rehearsal-free} setting, where data from earlier tasks cannot be revisited during later training, and task identities remain \textbf{unknown} at inference time.

\subsection{Initialization for the First Task}

For the first task $t$=$1$, PMA initializes a projector $\mathcal{P}^1$ together with two associated components: a lightweight Representation Descriptor $\mathcal{RD}^1$, implemented as an Multi-Layer Perceptron (MLP)-based autoencoder, and an initial router $\mathcal{R}^1$.
These components form the foundation for all subsequent progressive expansions.

For each training instance $x^{1,i} \in \mathcal{D}^1_{\text{train}}$, we first feed its image $x^{1,i}_{\text{img}}$ into the frozen visual encoder to obtain visual token embeddings, which are averaged to produce a global visual feature $\bm{x}^{1,i}_{\text{img}}\in\mathbb{R}^{d_1}$. 
The initial projector $\mathcal{P}^1$ maps this feature into the language embedding space:
\begin{equation}
    \bm{\tilde{x}}^{1,i}_{\text{img}} = w^1\mathcal{P}^1(\bm{x}^{1,i}_{\text{img}})\text{,}
\end{equation}
where $w^1$ represents the routing weight associated with $\mathcal{P}^1$. 
The router $\mathcal{R}^1$ computes this weight from a unified multimodal representation:
\begin{equation}
\begin{aligned}
    w^1 &= \text{Softmax}(\mathcal{R}^1(\bm{x}^{1,i}_{\text{fuse}})) \\
    \mathcal{R}^1(\bm{x}^{1,i}_{\text{fuse}}) &= {\bm{W}^1}^{T}\cdot\bm{x}^{1,i}_{\text{fuse}} 
\end{aligned}\text{,}\label{eq:router_fuse}
\end{equation}
where $\bm{W}^1$ is the learnable weight matrix of $\mathcal{R}^1$ and $\bm{x}^{1,i}_{\text{fuse}}\in\mathbb{R}^{d_1+d_2}=[\bm{x}^{1,i}_{\text{img}};\bm{x}^{1,i}_{\text{txt}}]$ concatenates the global visual feature with the averaged instruction-token embedding $\bm{x}^{1,i}_{\text{txt}}\in\mathbb{R}^{d_2}$. 
This fused representation enables routing decisions to depend on both visual content and instruction semantics, which is essential for MCIT. 
At $t$=$1$, PMA instantiates a default projector expert $\mathcal{P}^1$ together with the initial router $\mathcal{R}^1$. 
Since only a single expert is available, the matrix $\bm{W}^1$ has one column, and the Softmax degenerates to a constant selection, yielding $w^1$=$1.0$ for all samples.

$\mathcal{RD}^1$ operates on the same fused representation as $\mathcal{R}^1$, allowing it to model both the visual distribution and instruction semantics. 
Implemented as a small autoencoder, $\mathcal{RD}^1$ takes $\bm{x}^{1,i}_{\text{fuse}}$ as input and reconstructs it as $\bm{\hat{x}}^{1,i}_{\text{fuse}}$. 
The reconstruction error
\begin{equation}
    r = ||\bm{x}^{1,i}_{\text{fuse}} - \bm{\hat{x}}^{1,i}_{\text{fuse}}||^2_2 \label{eq:reconstruct}
\end{equation}
serves as a measure of how well the current projector configuration accounts for the new task.

After training $\mathcal{RD}^1$ on the first task, we compute the mean $\mu^1$ and standard deviation $\sigma^1$ of reconstruction errors across all training samples. 
These statistics define a reference distribution that characterizes the multimodal patterns of task $t$=$1$. 
For subsequent tasks, reconstruction errors produced by $\mathcal{RD}^1$ are compared against this baseline to determine whether incoming representations deviate substantially from those seen in the first task. 
This lightweight, data-driven mechanism allows PMA to detect task novelty and ensures that projector expansion is triggered only when necessary.

\subsection{Expansion for Subsequent Tasks} \label{sub_sec:expansion}

For each subsequent task $t\geq 2$, PMA determines whether the existing projector experts can adequately model the new multimodal representations. 
Given the fused representation $\bm{x}^{t,i}_{\text{fuse}}$ of an instance $i$ from task $t$, PMA evaluates it against \textbf{all previously instantiated descriptors} $\{\mathcal{RD}^j \mid 1 \le j \le t-1,\; \mathcal{RD}^j \text{ exists}\}$. 
Each $\mathcal{RD}^j$ produces reconstruction errors $r^{t,i}_{(j)}$, which are standardized using the statistics $(\mu^j,\sigma^j)$ collected from task $j$:
\begin{equation}
    z^{t,i}_{(j)} = \frac{r^{t,i}_{(j)}-\mu^j}{\sigma^j} \text{.}
\end{equation}
We rely on z-scores rather than raw reconstruction errors, as they normalize scale differences across descriptors and tasks, making deviations from prior distributions comparable and robust. 
Each sample thus obtains a z-score for each prior descriptor. 
For each $\mathcal{RD}^j$, PMA computes the proportion of samples $p_j$ whose z-scores satisfy $z^{t,i}_{(j)} \le \tau$.

If \textbf{all descriptors} yield proportions $p_j < 60\%$, PMA concludes that the new task introduces a multimodal distribution not captured by existing experts and allocates a new projector expert $\mathcal{P}^t$, a corresponding descriptor $\mathcal{RD}^t$, and a new weight column in $\mathcal{R}^t$. 
\textbf{Only} these newly added components are updated for task $t$, which biases the router toward assigning higher weights to the new projector on this task; 
we then compute the mean $\mu^t$ and standard deviation $\sigma^t$ of the reconstruction errors. 
All previously learned projectors, descriptors, and old router columns remain frozen.

Otherwise, PMA determines that no expansion is required and proceeds without introducing new components. 
In this case, PMA reuses all existing projector experts together with
their corresponding router weight columns. 
Crucially, during reuse, \textbf{all} projectors, descriptors, and router
columns \textbf{remain frozen} and are not updated using data from task
$t$, thereby preventing interference with previously learned tasks.
To identify the most relevant prior knowledge, PMA selects
$j^\star = \arg\max_j p_j$, corresponding to the most compatible prior task.
Since the projector $\mathcal{P}^{j^\star}$ was trained to receive
dominant routing weights on its originating task and the router parameters
remain frozen thereafter, and since both the descriptor and the router
operate on the same fused multimodal representation, the router is
encouraged to assign a larger routing weight to $\mathcal{P}^{j^\star}$
when processing a new task with a similar multimodal distribution, while
assigning smaller but non-zero weights to other projector experts.

This expansion mechanism promotes knowledge sharing across related tasks and ensures sub-linear parameter growth, as new experts are introduced only when PMA detects a genuinely novel multimodal distribution not explained by any previous RD. 
PMA thus provides a principled, data-driven mechanism for deciding when to reuse or expand projector capacity, enabling continual multimodal adaptation while maintaining stable visual-language alignment across tasks.

Throughout training, on both the first and subsequent tasks, the frozen pretrained projector $\mathcal{P}^0$ is kept as a stable alignment anchor, enabling PMA to preserve the core visual-language mapping established during pretraining while progressively adapting to new tasks. 
The final projected representation is given by
\begin{equation}
\bm{\tilde{x}}^{t,i}_{\text{img}}
= \frac{1}{1+t'}\Big(\mathcal{P}^0(\bm{x}^{t,i}_{\text{img}})
+ \sum_{j=1}^{t'} w^{j}\mathcal{P}^{j}(\bm{x}^{t,i}_{\text{img}})\Big)\text{,}\label{eq:map}
\end{equation}
where $t'$ is the number of instantiated projector experts ($1 \le t' < t$, ensuring sub-linear growth), and $w^{j}$ is the routing weight assigned to expert $j$ by the router $\mathcal{R}^t$, computed from the fused representation $\bm{x}^{t,i}_{\text{fuse}}$ via Equation (\ref{eq:router_fuse}). 
The resulting projected representation, aligned with the LLM's language embedding space, is fused with the instruction embeddings and passed to the LLM to generate the final prediction.

\subsection{Training Objective and Inference}

PMA optimizes two largely independent components: 
(1) the cross-modal alignment pathway, consisting of the projector experts and the router, and 
(2) the task-specific descriptor associated with each expert. 
The overall objective is
\begin{equation}
\mathcal{L} = \mathcal{L}_{\text{LM}} + \mathcal{L}_{\text{RD}},\label{eq:overall_loss}
\end{equation}
where $\mathcal{L}_{\text{LM}}$ is the autoregressive language modeling loss used for instruction tuning, and $\mathcal{L}_{\text{RD}}$ (Equation (\ref{eq:reconstruct})) is the reconstruction loss. 
These losses are fully decoupled: $\mathcal{L}_{\text{LM}}$ updates only the active projector expert and router column selected by PMA, along with the LLM-side PEFT parameters (depending on the combined MCIT method), whereas $\mathcal{L}_{\text{RD}}$ trains the descriptor alone and receives no gradient from the language modeling objective. 
This separation allows PMA to remain method-agnostic and integrate seamlessly with prior MCIT methods that mainly focus on mitigating CF in the LLM backbone.

\begin{table*}[t]
    \centering
    \caption{Main results of LLaVA-1.5-7B on UCIT. The middle columns for each task report performance after finetuning on the final task. The \textcolor{deepred}{\textbf{bold}} denotes the highest result. $*$ denotes results from our re-implementation; all other numbers are taken from MCITlib \cite{guo2025mcitlib}. HiDE and DISCO with \textbf{PMA} significantly outperform their corresponding vanilla methods.}
    \resizebox{1.0\linewidth}{!}{
    \begin{tabular}{>{\raggedright}p{2.0cm} *{11}{>{\centering\arraybackslash}p{1.2cm}}}
             \toprule[1.2pt]
             
        \multicolumn{1}{>{\raggedright}p{2.0cm}|}{Method} & \multicolumn{1}{>{\raggedright}p{1.7cm}|}{Venue} & \multicolumn{1}{c}{ImgNet-R} & \multicolumn{1}{c}{ArxivQA} & \multicolumn{1}{c}{VizWiz} & \multicolumn{1}{c}{IconQA} & \multicolumn{1}{c}{CLEVR} & \multicolumn{1}{c|}{Flickr30k} & \multicolumn{1}{c}{MFT~($\uparrow$)} & \multicolumn{1}{c}{MFN~($\uparrow$)} & \multicolumn{1}{c}{MAA~($\uparrow$)} & \multicolumn{1}{c}{BWT~($\uparrow$)}\\ 
        
        \midrule
        
         \multicolumn{1}{>{\raggedright}p{2.0cm}|}{Zero-shot} & \multicolumn{1}{>{\raggedright}p{1.7cm}|}{--} & \multicolumn{1}{c}{16.27} & \multicolumn{1}{c}{53.73} & \multicolumn{1}{c}{38.39} & \multicolumn{1}{c}{19.20} & \multicolumn{1}{c}{20.63} & \multicolumn{1}{c|}{41.88} & \multicolumn{1}{c}{--} & \multicolumn{1}{c}{31.68} & \multicolumn{1}{c}{--} & \multicolumn{1}{c}{--}\\ 
        
        \multicolumn{1}{>{\raggedright}p{2.0cm}|}{Individual} & \multicolumn{1}{>{\raggedright}p{1.7cm}|}{--} & \multicolumn{1}{c}{91.67} & \multicolumn{1}{c}{90.83} & \multicolumn{1}{c}{57.87} & \multicolumn{1}{c}{78.43} & \multicolumn{1}{c}{76.63} & \multicolumn{1}{c|}{61.72} & \multicolumn{1}{c}{--} & \multicolumn{1}{c}{76.19} & \multicolumn{1}{c}{--} & \multicolumn{1}{c}{--}\\ 
        
        \midrule\midrule
        
        \multicolumn{1}{>{\raggedright}p{2.0cm}|}{LoRA-FT} & \multicolumn{1}{>{\raggedright}p{1.7cm}|}{ICLR'22} & \multicolumn{1}{c}{58.03} & \multicolumn{1}{c}{77.63} & \multicolumn{1}{c}{44.39} & \multicolumn{1}{c}{67.40} & \multicolumn{1}{c}{61.77} & \multicolumn{1}{c|}{{58.22}} & \multicolumn{1}{c}{{76.89}} & \multicolumn{1}{c}{61.24} & \multicolumn{1}{c}{76.55} & \multicolumn{1}{c}{-18.78}\\ 
        
        \multicolumn{1}{>{\raggedright}p{2.0cm}|}{OLoRA} & \multicolumn{1}{>{\raggedright}p{1.7cm}|}{EMNLP'23} & \multicolumn{1}{c}{77.50} & \multicolumn{1}{c}{78.07} & \multicolumn{1}{c}{44.50} & \multicolumn{1}{c}{63.13} & \multicolumn{1}{c}{64.73} & \multicolumn{1}{c|}{58.16} & \multicolumn{1}{c}{76.01} & \multicolumn{1}{c}{64.35} & \multicolumn{1}{c}{78.02} & \multicolumn{1}{c}{-13.99}\\ 
        
        \multicolumn{1}{>{\raggedright}p{2.0cm}|}{MoELoRA} & \multicolumn{1}{>{\raggedright}p{1.7cm}|}{NeurIPS'24}  & \multicolumn{1}{c}{70.07} & \multicolumn{1}{c}{77.70} & \multicolumn{1}{c}{44.69} & \multicolumn{1}{c}{50.03} & \multicolumn{1}{c}{54.03} & \multicolumn{1}{c|}{57.34} & \multicolumn{1}{c}{71.17} & \multicolumn{1}{c}{58.98} & \multicolumn{1}{c}{75.08} & \multicolumn{1}{c}{-14.63}\\

        \multicolumn{1}{>{\raggedright}p{2.0cm}|}{CL-MoE} & \multicolumn{1}{>{\raggedright}p{1.7cm}|}{CVPR'25} & \multicolumn{1}{c}{66.33} & \multicolumn{1}{c}{77.00} & \multicolumn{1}{c}{44.78} & \multicolumn{1}{c}{51.87} & \multicolumn{1}{c}{53.53} & \multicolumn{1}{c|}{57.42} & \multicolumn{1}{c}{71.46} & \multicolumn{1}{c}{58.49} & \multicolumn{1}{c}{74.19} & \multicolumn{1}{c}{-15.56}\\ 

        \multicolumn{1}{>{\raggedright}p{2.0cm}|}{SEFE} & \multicolumn{1}{>{\raggedright}p{1.7cm}|}{ICML'25} & \multicolumn{1}{c}{80.83} & \multicolumn{1}{c}{78.00} & \multicolumn{1}{c}{47.01} & \multicolumn{1}{c}{{69.63}} & \multicolumn{1}{c}{{65.83}} & \multicolumn{1}{c|}{57.92} & \multicolumn{1}{c}{75.98} & \multicolumn{1}{c}{66.54} & \multicolumn{1}{c}{78.76} & \multicolumn{1}{c}{-11.33}\\ 

   \hline

        \multicolumn{1}{>{\raggedright}p{2.0cm}|}{HiDE} & \multicolumn{1}{>{\raggedright}p{1.7cm}|}{ACL'25} & \multicolumn{1}{c}{84.03} & \multicolumn{1}{c}{90.73} & \multicolumn{1}{c}{44.43} & \multicolumn{1}{c}{58.93} & \multicolumn{1}{c}{41.37} & \multicolumn{1}{c|}{54.25} & \multicolumn{1}{c}{69.96} & \multicolumn{1}{c}{62.29} & \multicolumn{1}{c}{77.32} & \multicolumn{1}{c}{-9.20}\\

        \multicolumn{1}{>{\raggedright}p{2.0cm}|}{HiDE*} & \multicolumn{1}{>{\raggedright}p{1.7cm}|}{ACL'25} & \multicolumn{1}{c}{86.00} & \multicolumn{1}{c}{90.60} & \multicolumn{1}{c}{45.33} & \multicolumn{1}{c}{66.13} & \multicolumn{1}{c}{49.03} & \multicolumn{1}{c|}{52.30} & \multicolumn{1}{c}{70.24} & \multicolumn{1}{c}{64.90} & \multicolumn{1}{c}{78.61} & \multicolumn{1}{c}{-5.34}\\ 

       \rowcolor{gray!20} \multicolumn{1}{>{\raggedright}p{2.0cm}|}{\textbf{\ \ \ + PMA}} & \multicolumn{1}{>{\raggedright}p{1.7cm}|}{\textbf{Ours}} & \multicolumn{1}{c}{84.77} & \multicolumn{1}{c}{93.83} & \multicolumn{1}{c}{50.35} & \multicolumn{1}{c}{70.67} & \multicolumn{1}{c}{53.07} & \multicolumn{1}{c|}{54.55} & \multicolumn{1}{c}{72.50} & \multicolumn{1}{c}{67.87} & \multicolumn{1}{c}{80.57} & \multicolumn{1}{c}{-4.63}\\

   \hline

        \multicolumn{1}{>{\raggedright}p{2.0cm}|}{DISCO} & \multicolumn{1}{>{\raggedright}p{1.7cm}|}{ICCV'25} & \multicolumn{1}{c}{{87.43}} & \multicolumn{1}{c}{{93.07}} & \multicolumn{1}{c}{46.96} & \multicolumn{1}{c}{68.13} & \multicolumn{1}{c}{65.70} & \multicolumn{1}{c|}{56.69} & \multicolumn{1}{c}{75.87} & \multicolumn{1}{c}{{69.66}} & \multicolumn{1}{c}{{81.60}} & \multicolumn{1}{c}{-7.45}\\ 

        \multicolumn{1}{>{\raggedright}p{2.0cm}|}{DISCO*} & \multicolumn{1}{>{\raggedright}p{1.7cm}|}{ICCV'25} & \multicolumn{1}{c}{88.13} & \multicolumn{1}{c}{95.00} & \multicolumn{1}{c}{46.65} & \multicolumn{1}{c}{71.50} & \multicolumn{1}{c}{53.33} & \multicolumn{1}{c|}{56.02} & \multicolumn{1}{c}{75.20} & \multicolumn{1}{c}{68.44} & \multicolumn{1}{c}{81.36} & \multicolumn{1}{c}{-6.76}\\ 
        
       \rowcolor{gray!20} \multicolumn{1}{>{\raggedright}p{2.0cm}|}{\textbf{\ \ \ + PMA}} & \multicolumn{1}{>{\raggedright}p{1.7cm}|}{\textbf{Ours}} & \multicolumn{1}{c}{88.00} & \multicolumn{1}{c}{95.50} & \multicolumn{1}{c}{52.66} & \multicolumn{1}{c}{74.50} & \multicolumn{1}{c}{69.77} & \multicolumn{1}{c|}{58.83} & \multicolumn{1}{c}{\textcolor{deepred}{\textbf{77.57}}} & \multicolumn{1}{c}{\textcolor{deepred}{\textbf{73.21}}} & \multicolumn{1}{c}{\textcolor{deepred}{\textbf{83.98}}} & \multicolumn{1}{c}{\textcolor{deepred}{\textbf{-4.36}}}\\

        \bottomrule[1.2pt]
    \end{tabular}
    }
    \label{tab:llava_ucit}
\end{table*}

\begin{table*}[t]
    \centering
        \caption{Main results of the InternVL-Chat-7B model on the UCIT benchmark.}
    \resizebox{1.0\linewidth}{!}{
    \begin{tabular}{>{\raggedright}p{2.0cm} *{11}{>{\centering\arraybackslash}p{1.2cm}}}
             \toprule[1.2pt]
             
        \multicolumn{1}{>{\raggedright}p{2.0cm}|}{Method} & \multicolumn{1}{>{\raggedright}p{1.7cm}|}{Venue} & \multicolumn{1}{c}{ImgNet-R} & \multicolumn{1}{c}{ArxivQA} & \multicolumn{1}{c}{VizWiz} & \multicolumn{1}{c}{IconQA} & \multicolumn{1}{c}{CLEVR} & \multicolumn{1}{c|}{Flickr30k} & \multicolumn{1}{c}{MFT~($\uparrow$)} & \multicolumn{1}{c}{MFN~($\uparrow$)} & \multicolumn{1}{c}{MAA~($\uparrow$)} & \multicolumn{1}{c}{BWT~($\uparrow$)}\\ 
        
        \midrule
        
         \multicolumn{1}{>{\raggedright}p{2.0cm}|}{Zero-shot} & \multicolumn{1}{>{\raggedright}p{1.7cm}|}{--} & \multicolumn{1}{c}{21.10} & \multicolumn{1}{c}{63.20} & \multicolumn{1}{c}{40.59} & \multicolumn{1}{c}{24.70} & \multicolumn{1}{c}{21.20} & \multicolumn{1}{c|}{44.67} & \multicolumn{1}{c}{--} & \multicolumn{1}{c}{35.91} & \multicolumn{1}{c}{--} & \multicolumn{1}{c}{--}\\

   \multicolumn{1}{>{\raggedright}p{2.0cm}|}{Individual} & \multicolumn{1}{>{\raggedright}p{1.7cm}|}{--} & \multicolumn{1}{c}{95.40} & \multicolumn{1}{c}{92.70} & \multicolumn{1}{c}{62.53} & \multicolumn{1}{c}{82.87} & \multicolumn{1}{c}{83.80} & \multicolumn{1}{c|}{58.82} & \multicolumn{1}{c}{--} & \multicolumn{1}{c}{79.35} & \multicolumn{1}{c}{--} & \multicolumn{1}{c}{--}\\

        \midrule\midrule

        \multicolumn{1}{>{\raggedright}p{2.0cm}|}{LoRA-FT} & \multicolumn{1}{>{\raggedright}p{1.7cm}|}{ICLR'22} & \multicolumn{1}{c}{75.90} & \multicolumn{1}{c}{77.60} & \multicolumn{1}{c}{44.57} & \multicolumn{1}{c}{68.37} & \multicolumn{1}{c}{69.20} & \multicolumn{1}{c|}{{58.03}} & \multicolumn{1}{c}{{73.95}} & \multicolumn{1}{c}{65.61} & \multicolumn{1}{c}{78.62} & \multicolumn{1}{c}{-8.34}\\

        \multicolumn{1}{>{\raggedright}p{2.0cm}|}{OLoRA} & \multicolumn{1}{>{\raggedright}p{1.7cm}|}{EMNLP'23} & \multicolumn{1}{c}{86.37} & \multicolumn{1}{c}{93.73} & \multicolumn{1}{c}{44.13} & \multicolumn{1}{c}{68.10} & \multicolumn{1}{c}{64.53} & \multicolumn{1}{c|}{{56.19}} & \multicolumn{1}{c}{{72.36}} & \multicolumn{1}{c}{68.84} & \multicolumn{1}{c}{81.10} & \multicolumn{1}{c}{-3.51}\\  

   \multicolumn{1}{>{\raggedright}p{2.0cm}|}{MoELoRA} & \multicolumn{1}{>{\raggedright}p{1.7cm}|}{NeurIPS'24}  & \multicolumn{1}{c}{72.03} & \multicolumn{1}{c}{78.00} & \multicolumn{1}{c}{44.82} & \multicolumn{1}{c}{68.83} & \multicolumn{1}{c}{65.87} & \multicolumn{1}{c|}{{57.67}} & \multicolumn{1}{c}{{73.04}} & \multicolumn{1}{c}{64.54} & \multicolumn{1}{c}{77.78} & \multicolumn{1}{c}{-8.51}\\

        \multicolumn{1}{>{\raggedright}p{2.0cm}|}{CL-MoE} & \multicolumn{1}{>{\raggedright}p{1.7cm}|}{CVPR'25} & \multicolumn{1}{c}{74.23} & \multicolumn{1}{c}{78.77} & \multicolumn{1}{c}{44.77} & \multicolumn{1}{c}{53.90} & \multicolumn{1}{c}{72.73} & \multicolumn{1}{c|}{58.23} & \multicolumn{1}{c}{71.39} & \multicolumn{1}{c}{63.77} & \multicolumn{1}{c}{77.44} & \multicolumn{1}{c}{-9.15}\\

        \multicolumn{1}{>{\raggedright}p{2.0cm}|}{SEFE} & \multicolumn{1}{>{\raggedright}p{1.7cm}|}{ICML'25} & \multicolumn{1}{c}{85.93} & \multicolumn{1}{c}{76.90} & \multicolumn{1}{c}{47.42} & \multicolumn{1}{c}{69.77} & \multicolumn{1}{c}{63.37} & \multicolumn{1}{c|}{{58.19}} & \multicolumn{1}{c}{{73.55}} & \multicolumn{1}{c}{66.93} & \multicolumn{1}{c}{79.44} & \multicolumn{1}{c}{-7.94}\\ 

   \hline

        \multicolumn{1}{>{\raggedright}p{2.0cm}|}{HiDE} & \multicolumn{1}{>{\raggedright}p{1.7cm}|}{ACL'25} & \multicolumn{1}{c}{90.00} & \multicolumn{1}{c}{93.77} & \multicolumn{1}{c}{50.49} & \multicolumn{1}{c}{71.03} & \multicolumn{1}{c}{61.10} & \multicolumn{1}{c|}{{55.08}} & \multicolumn{1}{c}{{76.29}} & \multicolumn{1}{c}{70.25} & \multicolumn{1}{c}{82.97} & \multicolumn{1}{c}{-7.25}\\

        \multicolumn{1}{>{\raggedright}p{2.0cm}|}{HiDE*} & \multicolumn{1}{>{\raggedright}p{1.7cm}|}{ACL'25}  & \multicolumn{1}{c}{88.73} & \multicolumn{1}{c}{91.83} & \multicolumn{1}{c}{46.69} & \multicolumn{1}{c}{61.03} & \multicolumn{1}{c}{66.23} & \multicolumn{1}{c|}{54.68} & \multicolumn{1}{c}{73.07} & \multicolumn{1}{c}{68.20} & \multicolumn{1}{c}{81.74} & \multicolumn{1}{c}{-4.88}\\ 

            \rowcolor{gray!20}  \multicolumn{1}{>{\raggedright}p{2.0cm}|}{\textbf{\ \ \ + PMA}} & \multicolumn{1}{>{\raggedright}p{1.7cm}|}{\textbf{Ours}} & \multicolumn{1}{c}{91.70} & \multicolumn{1}{c}{94.87} & \multicolumn{1}{c}{51.66} & \multicolumn{1}{c}{71.00} & \multicolumn{1}{c}{69.20} & \multicolumn{1}{c|}{56.88} & \multicolumn{1}{c}{76.91} & \multicolumn{1}{c}{72.55} & \multicolumn{1}{c}{84.28} & \multicolumn{1}{c}{-4.36}\\

   \hline
    
        \multicolumn{1}{>{\raggedright}p{2.0cm}|}{DISCO} & \multicolumn{1}{>{\raggedright}p{1.7cm}|}{ICCV'25} & \multicolumn{1}{c}{92.13} & \multicolumn{1}{c}{94.13} & \multicolumn{1}{c}{48.11} & \multicolumn{1}{c}{73.90} & \multicolumn{1}{c}{67.53} & \multicolumn{1}{c|}{{58.06}} & \multicolumn{1}{c}{{78.92}} & \multicolumn{1}{c}{72.31} & \multicolumn{1}{c}{84.24} & \multicolumn{1}{c}{-7.93}\\ 

        \multicolumn{1}{>{\raggedright}p{2.0cm}|}{DISCO*} & \multicolumn{1}{>{\raggedright}p{1.7cm}|}{ICCV'25} & \multicolumn{1}{c}{91.17} & \multicolumn{1}{c}{94.87} & \multicolumn{1}{c}{47.02} & \multicolumn{1}{c}{69.37} & \multicolumn{1}{c}{71.83} & \multicolumn{1}{c|}{{57.08}} & \multicolumn{1}{c}{{73.70}} & \multicolumn{1}{c}{71.89} & \multicolumn{1}{c}{82.85} & \multicolumn{1}{c}{-1.81}\\

           \rowcolor{gray!20}  \multicolumn{1}{>{\raggedright}p{2.0cm}|}{\textbf{\ \ \ + PMA}} & \multicolumn{1}{>{\raggedright}p{1.7cm}|}{\textbf{Ours}} & \multicolumn{1}{c}{93.37} & \multicolumn{1}{c}{95.07} & \multicolumn{1}{c}{59.90} & \multicolumn{1}{c}{74.97} & \multicolumn{1}{c}{82.83} & \multicolumn{1}{c|}{58.40} & \multicolumn{1}{c}{\textcolor{deepred}{\textbf{79.04}}} & \multicolumn{1}{c}{\textcolor{deepred}{\textbf{77.42}}}  & \multicolumn{1}{c}{\textcolor{deepred}{\textbf{86.59}}} & \multicolumn{1}{c}{\textcolor{deepred}{\textbf{-1.61}}}\\ 

        \bottomrule[1.2pt]
    \end{tabular}
    }
    \label{tab:internvl_ucit}
\end{table*}

\section{Experimental Settings} \label{sec:exp_setting}

\subsection{Datasets and Benchmarks} 
As the MLLMs have already seen large-scale image-text pairs during pretraining, we adopt two benchmarks from MCITlib \cite{guo2025mcitlib} designed to mitigate information leakage in MCIT training: 
(1) \textbf{UCIT} benchmark \cite{DBLP:conf/acl/GuoZXZWZL25} consists of ImageNet-R (ImgNet-R) \cite{hendrycks2021many}, ArxivQA \cite{li2024multimodal}, VizWiz-Caption (VizWiz) \cite{gurari2018vizwiz}, IconQA \cite{lu2021iconqa}, CLEVR-Math (CLEVR) \cite{DBLP:conf/nesy/LindstromA22}, and Flickr30k \cite{plummer2015flickr30k}, including image captioning, Visual Question Answering (VQA), and multiple-choice reasoning tasks. 
The MLLMs exhibit weak zero-shot performance, suggesting a low risk of information leakage. 
All datasets are trained in the above order as in the main experiments. 
(2) \textbf{MLLM-DCL} benchmark \cite{zhao2025mllm} extends to downstream tasks from five domains---Remote Sensing (RS), Medicine (Med), Autonomous Driving (AD), Science (Sci), and Finance (Fin)---trained in the order RS $\rightarrow$ Med $\rightarrow$ AD $\rightarrow$ Sci $\rightarrow$ Fin, incorporating RSVQA \cite{lobry2020rsvqa}, PathVQA \cite{he2020pathvqa}, DriveLM \cite{sima2024drivelm}, AI2D \cite{kembhavi2016diagram}, Sciverse \cite{DBLP:conf/acl/GuoZ0GJWH25}, MapQA \cite{chang2022mapqa}, TQA \cite{kembhavi2017you}, and FinVis \cite{wang2023finvis} datasets.

\begin{table*}[t]
    \centering
    \caption{Main results of the LLaVA-1.5-7B model on the MLLM-DCL benchmark.}
    \resizebox{0.95\linewidth}{!}{
    \begin{tabular}{>{\raggedright}p{2.0cm} *{10}{>{\centering\arraybackslash}p{1.2cm}}}
        \toprule[1.2pt]
        
        \multicolumn{1}{>{\raggedright}p{2.0cm}|}{Method} & \multicolumn{1}{>{\raggedright}p{1.7cm}|}{Venue} & \multicolumn{1}{c}{RS} & \multicolumn{1}{c}{Med} & \multicolumn{1}{c}{AD} & \multicolumn{1}{c}{Sci} & \multicolumn{1}{c|}{Fin} & \multicolumn{1}{c}{MFT~($\uparrow$)} & \multicolumn{1}{c}{MFN~($\uparrow$)} & \multicolumn{1}{c}{MAA~($\uparrow$)} & \multicolumn{1}{c}{BWT~($\uparrow$)}\\ 
        
        \midrule
        
        \multicolumn{1}{>{\raggedright}p{2.0cm}|}{Zero-shot} & \multicolumn{1}{>{\raggedright}p{1.7cm}|}{--} & \multicolumn{1}{c}{32.29} & \multicolumn{1}{c}{28.28} & \multicolumn{1}{c}{15.59} & \multicolumn{1}{c}{35.55} & \multicolumn{1}{c|}{62.56} & \multicolumn{1}{c}{--} & \multicolumn{1}{c}{34.85} & \multicolumn{1}{c}{--} & \multicolumn{1}{c}{--}\\ 
        
        \multicolumn{1}{>{\raggedright}p{2.0cm}|}{Individual} & \multicolumn{1}{>{\raggedright}p{1.7cm}|}{--} & \multicolumn{1}{c}{78.15} & \multicolumn{1}{c}{58.20} & \multicolumn{1}{c}{52.77} & \multicolumn{1}{c}{49.32} & \multicolumn{1}{c|}{88.02} & \multicolumn{1}{c}{--} & \multicolumn{1}{c}{65.29} & \multicolumn{1}{c}{--} & \multicolumn{1}{c}{--}\\ 
        
        \midrule\midrule
        
        \multicolumn{1}{>{\raggedright}p{2.0cm}|}{LoRA-FT} & \multicolumn{1}{>{\raggedright}p{1.7cm}|}{ICLR'22} & \multicolumn{1}{c}{69.65} & \multicolumn{1}{c}{41.59} & \multicolumn{1}{c}{25.43} & \multicolumn{1}{c}{40.88} & \multicolumn{1}{c|}{87.45} & \multicolumn{1}{c}{64.98} & \multicolumn{1}{c}{53.00} & \multicolumn{1}{c}{61.13} & \multicolumn{1}{c}{-14.97}\\ 
        
        \multicolumn{1}{>{\raggedright}p{2.0cm}|}{OLoRA} & \multicolumn{1}{>{\raggedright}p{1.7cm}|}{EMNLP'23} & \multicolumn{1}{c}{74.64} & \multicolumn{1}{c}{44.42} & \multicolumn{1}{c}{30.02} & \multicolumn{1}{c}{41.47} & \multicolumn{1}{c|}{87.15} & \multicolumn{1}{c}{65.16} & \multicolumn{1}{c}{55.54} & \multicolumn{1}{c}{62.12} & \multicolumn{1}{c}{-12.03}\\ 
        
        \multicolumn{1}{>{\raggedright}p{2.0cm}|}{MoELoRA} & \multicolumn{1}{>{\raggedright}p{1.7cm}|}{NeurIPS'24}  & \multicolumn{1}{c}{77.54} & \multicolumn{1}{c}{41.85} & \multicolumn{1}{c}{27.62} & \multicolumn{1}{c}{40.13} & \multicolumn{1}{c|}{86.75} & \multicolumn{1}{c}{64.94} & \multicolumn{1}{c}{54.78} & \multicolumn{1}{c}{61.76} & \multicolumn{1}{c}{-12.71}\\

        \multicolumn{1}{>{\raggedright}p{2.0cm}|}{CL-MoE} & \multicolumn{1}{>{\raggedright}p{1.7cm}|}{CVPR'25} & \multicolumn{1}{c}{71.34} & \multicolumn{1}{c}{46.84} & \multicolumn{1}{c}{26.33} & \multicolumn{1}{c}{41.17} & \multicolumn{1}{c|}{88.74} & \multicolumn{1}{c}{{66.06}} & \multicolumn{1}{c}{54.88} & \multicolumn{1}{c}{61.79} & \multicolumn{1}{c}{-13.97}\\ 

        \multicolumn{1}{>{\raggedright}p{2.0cm}|}{SEFE} & \multicolumn{1}{>{\raggedright}p{1.7cm}|}{ICML'25} & \multicolumn{1}{c}{77.26} & \multicolumn{1}{c}{50.37} & \multicolumn{1}{c}{37.21} & \multicolumn{1}{c}{40.87} & \multicolumn{1}{c|}{86.82} & \multicolumn{1}{c}{65.01} & \multicolumn{1}{c}{58.51} & \multicolumn{1}{c}{63.63} & \multicolumn{1}{c}{-8.13}\\ 

        \hline

        \multicolumn{1}{>{\raggedright}p{2.0cm}|}{HiDE} & \multicolumn{1}{>{\raggedright}p{1.7cm}|}{ACL'25} & \multicolumn{1}{c}{74.31} & \multicolumn{1}{c}{48.95} & \multicolumn{1}{c}{33.21} & \multicolumn{1}{c}{38.54} & \multicolumn{1}{c|}{81.55} & \multicolumn{1}{c}{60.77} & \multicolumn{1}{c}{55.31} & \multicolumn{1}{c}{60.68} & \multicolumn{1}{c}{-6.82}\\ 

       \multicolumn{1}{>{\raggedright}p{2.0cm}|}{HiDE*} & \multicolumn{1}{>{\raggedright}p{1.7cm}|}{ACL'25} & \multicolumn{1}{c}{74.68} & \multicolumn{1}{c}{50.37} & \multicolumn{1}{c}{34.14} & \multicolumn{1}{c}{40.14} & \multicolumn{1}{c|}{80.88} & \multicolumn{1}{c}{61.77} & \multicolumn{1}{c}{56.04} & \multicolumn{1}{c}{62.30} & \multicolumn{1}{c}{-5.73}\\ 

         \rowcolor{gray!20}  \multicolumn{1}{>{\raggedright}p{2.0cm}|}{\ \ \ \textbf{+ PMA}} & \multicolumn{1}{>{\raggedright}p{1.7cm}|}{\textbf{Ours}} & \multicolumn{1}{c}{75.94} & \multicolumn{1}{c}{52.38} & \multicolumn{1}{c}{35.44} & \multicolumn{1}{c}{42.46} & \multicolumn{1}{c|}{80.68} & \multicolumn{1}{c}{62.67} & \multicolumn{1}{c}{57.38} & \multicolumn{1}{c}{63.49} & \multicolumn{1}{c}{-5.29}\\ 

        \hline

        \multicolumn{1}{>{\raggedright}p{2.0cm}|}{DISCO} & \multicolumn{1}{>{\raggedright}p{1.7cm}|}{ICCV'25} & \multicolumn{1}{c}{76.49} & \multicolumn{1}{c}{44.48} & \multicolumn{1}{c}{44.84} & \multicolumn{1}{c}{{46.61}} & \multicolumn{1}{c|}{{89.22}} & \multicolumn{1}{c}{64.78} & \multicolumn{1}{c}{60.33} & \multicolumn{1}{c}{63.93} & \multicolumn{1}{c}{-5.57}\\ 

       \multicolumn{1}{>{\raggedright}p{2.0cm}|}{DISCO*} & \multicolumn{1}{>{\raggedright}p{1.7cm}|}{ICCV'25} & \multicolumn{1}{c}{73.61} & \multicolumn{1}{c}{44.86} & \multicolumn{1}{c}{47.92} & \multicolumn{1}{c}{44.83} & \multicolumn{1}{c|}{ 84.96} & \multicolumn{1}{c}{64.61} & \multicolumn{1}{c}{59.24} & \multicolumn{1}{c}{64.01} & \multicolumn{1}{c}{-5.37}\\ 

       \rowcolor{gray!20} \multicolumn{1}{>{\raggedright}p{2.0cm}|}{\ \ \ \textbf{+ PMA}} & \multicolumn{1}{>{\raggedright}p{1.7cm}|}{\textbf{Ours}} & \multicolumn{1}{c}{76.69} & \multicolumn{1}{c}{44.25} & \multicolumn{1}{c}{52.64} & \multicolumn{1}{c}{49.42} & \multicolumn{1}{c|}{89.83} & \multicolumn{1}{c}{\textcolor{deepred}{\textbf{66.28}}} & \multicolumn{1}{c}{\textcolor{deepred}{\textbf{62.57}}} & \multicolumn{1}{c}{\textcolor{deepred}{\textbf{65.38}}} & \multicolumn{1}{c}{\textcolor{deepred}{\textbf{-3.71}}}\\

        \bottomrule[1.2pt]
    \end{tabular}}

    \label{tab:llava_dcl}
\end{table*}

\begin{table*}[t]
    \centering
    \caption{Main results of the InternVL-Chat-7B model on the MLLM-DCL benchmark.}
    \resizebox{0.95\linewidth}{!}{
    \begin{tabular}{>{\raggedright}p{2.0cm} *{10}{>{\centering\arraybackslash}p{1.2cm}}}
        \toprule[1.2pt]
        
        \multicolumn{1}{>{\raggedright}p{2.0cm}|}{Method} & \multicolumn{1}{>{\raggedright}p{1.7cm}|}{Venue} & \multicolumn{1}{c}{RS} & \multicolumn{1}{c}{Med} & \multicolumn{1}{c}{AD} & \multicolumn{1}{c}{Sci} & \multicolumn{1}{c|}{Fin} & \multicolumn{1}{c}{MFT~($\uparrow$)} & \multicolumn{1}{c}{MFN~($\uparrow$)} & \multicolumn{1}{c}{MAA~($\uparrow$)} & \multicolumn{1}{c}{BWT~($\uparrow$)}\\ 
        
        \midrule
        
        \multicolumn{1}{>{\raggedright}p{2.0cm}|}{Zero-shot} & \multicolumn{1}{>{\raggedright}p{1.7cm}|}{--} & \multicolumn{1}{c}{31.16} & \multicolumn{1}{c}{29.81} & \multicolumn{1}{c}{14.06} & \multicolumn{1}{c}{33.93} & \multicolumn{1}{c|}{64.32} & \multicolumn{1}{c}{--} & \multicolumn{1}{c}{34.66} & \multicolumn{1}{c}{--} & \multicolumn{1}{c}{--}\\ 
        
        \multicolumn{1}{>{\raggedright}p{2.0cm}|}{Individual} & \multicolumn{1}{>{\raggedright}p{1.7cm}|}{--} & \multicolumn{1}{c}{81.49} & \multicolumn{1}{c}{66.42} & \multicolumn{1}{c}{54.56} & \multicolumn{1}{c}{54.48} & \multicolumn{1}{c|}{91.24} & \multicolumn{1}{c}{--} & \multicolumn{1}{c}{69.64} & \multicolumn{1}{c}{--} & \multicolumn{1}{c}{--}\\  
        
        \midrule\midrule

        \multicolumn{1}{>{\raggedright}p{2.0cm}|}{LoRA-FT} & \multicolumn{1}{>{\raggedright}p{1.7cm}|}{ICLR'22} & \multicolumn{1}{c}{69.93} & \multicolumn{1}{c}{52.17} & \multicolumn{1}{c}{33.04} & \multicolumn{1}{c}{42.67} & \multicolumn{1}{c|}{91.07} & \multicolumn{1}{c}{69.06} & \multicolumn{1}{c}{57.78} & \multicolumn{1}{c}{65.22} & \multicolumn{1}{c}{-14.11}\\

        \multicolumn{1}{>{\raggedright}p{2.0cm}|}{OLoRA} & \multicolumn{1}{>{\raggedright}p{1.7cm}|}{EMNLP'23} & \multicolumn{1}{c}{74.48} & \multicolumn{1}{c}{54.16} & \multicolumn{1}{c}{39.60} & \multicolumn{1}{c}{48.30} & \multicolumn{1}{c|}{88.54} & \multicolumn{1}{c}{65.51} & \multicolumn{1}{c}{61.02} & \multicolumn{1}{c}{65.83} & \multicolumn{1}{c}{-5.62}\\

        \multicolumn{1}{>{\raggedright}p{2.0cm}|}{MoELoRA} & \multicolumn{1}{>{\raggedright}p{1.7cm}|}{NeurIPS'24}  & \multicolumn{1}{c}{69.90} & \multicolumn{1}{c}{52.08} & \multicolumn{1}{c}{33.17} & \multicolumn{1}{c}{42.19} & \multicolumn{1}{c|}{90.58} & \multicolumn{1}{c}{68.83} & \multicolumn{1}{c}{57.58} & \multicolumn{1}{c}{65.97} & \multicolumn{1}{c}{-14.06}\\

        \multicolumn{1}{>{\raggedright}p{2.0cm}|}{CL-MoE} & \multicolumn{1}{>{\raggedright}p{1.7cm}|}{CVPR'25} & \multicolumn{1}{c}{78.12} & \multicolumn{1}{c}{52.51} & \multicolumn{1}{c}{35.53} & \multicolumn{1}{c}{42.69} & \multicolumn{1}{c|}{91.24} & \multicolumn{1}{c}{{69.22}} & \multicolumn{1}{c}{60.02} & \multicolumn{1}{c}{67.60} & \multicolumn{1}{c}{-11.51}\\ 

        \multicolumn{1}{>{\raggedright}p{2.0cm}|}{SEFE} & \multicolumn{1}{>{\raggedright}p{1.7cm}|}{ICML'25} & \multicolumn{1}{c}{78.21} & \multicolumn{1}{c}{57.59} & \multicolumn{1}{c}{51.45} & \multicolumn{1}{c}{44.65} & \multicolumn{1}{c|}{91.37} & \multicolumn{1}{c}{69.55} & \multicolumn{1}{c}{64.65} & \multicolumn{1}{c}{68.84} & \multicolumn{1}{c}{-6.12}\\ 

        \hline

      \multicolumn{1}{>{\raggedright}p{2.0cm}|}{HiDE} & \multicolumn{1}{>{\raggedright}p{1.7cm}|}{ACL'25} & \multicolumn{1}{c}{75.40} & \multicolumn{1}{c}{57.66} & \multicolumn{1}{c}{36.73} & \multicolumn{1}{c}{41.48} & \multicolumn{1}{c|}{88.59} & \multicolumn{1}{c}{65.26} & \multicolumn{1}{c}{59.97} & \multicolumn{1}{c}{65.94} & \multicolumn{1}{c}{-6.61}\\ 

        \multicolumn{1}{>{\raggedright}p{2.0cm}|}{HiDE*} & \multicolumn{1}{>{\raggedright}p{1.7cm}|}{ACL'25} & \multicolumn{1}{c}{78.54} & \multicolumn{1}{c}{57.64} & \multicolumn{1}{c}{37.68} & \multicolumn{1}{c}{48.45} & \multicolumn{1}{c|}{90.46} & \multicolumn{1}{c}{66.47} & \multicolumn{1}{c}{62.55} & \multicolumn{1}{c}{67.24} & \multicolumn{1}{c}{-3.92}\\

      \rowcolor{gray!20}  \multicolumn{1}{>{\raggedright}p{2.0cm}|}{\ \ \ \textbf{+ PMA}} & \multicolumn{1}{>{\raggedright}p{1.7cm}|}{\textbf{Ours}} & \multicolumn{1}{c}{81.05} & \multicolumn{1}{c}{60.17} & \multicolumn{1}{c}{40.14} & \multicolumn{1}{c}{50.92} & \multicolumn{1}{c|}{92.95} & \multicolumn{1}{c}{68.60} & \multicolumn{1}{c}{65.05} & \multicolumn{1}{c}{69.34} & \multicolumn{1}{c}{-3.55}\\

        \hline

        \multicolumn{1}{>{\raggedright}p{2.0cm}|}{DISCO} & \multicolumn{1}{>{\raggedright}p{1.7cm}|}{ICCV'25} & \multicolumn{1}{c}{77.90} & \multicolumn{1}{c}{47.50} & \multicolumn{1}{c}{49.13} & \multicolumn{1}{c}{{49.37}} & \multicolumn{1}{c|}{{90.92}} & \multicolumn{1}{c}{68.55} & \multicolumn{1}{c}{62.96} & \multicolumn{1}{c}{67.81} & \multicolumn{1}{c}{-6.98}\\ 

        \multicolumn{1}{>{\raggedright}p{2.0cm}|}{DISCO*} & \multicolumn{1}{>{\raggedright}p{1.7cm}|}{ICCV'25} & \multicolumn{1}{c}{77.70} & \multicolumn{1}{c}{50.19} & \multicolumn{1}{c}{53.41} & \multicolumn{1}{c}{{50.30}} & \multicolumn{1}{c|}{{90.67}} & \multicolumn{1}{c}{68.69} & \multicolumn{1}{c}{64.45} & \multicolumn{1}{c}{68.16} & \multicolumn{1}{c}{-4.24}\\

       \rowcolor{gray!20} \multicolumn{1}{>{\raggedright}p{2.0cm}|}{\ \ \ \textbf{+ PMA}} & \multicolumn{1}{>{\raggedright}p{1.7cm}|}{\textbf{Ours}} & \multicolumn{1}{c}{81.20} & \multicolumn{1}{c}{56.63} & \multicolumn{1}{c}{53.07} & \multicolumn{1}{c}{52.70} & \multicolumn{1}{c|}{92.66} & \multicolumn{1}{c}{\textcolor{deepred}{\textbf{69.87}}} & \multicolumn{1}{c}{\textcolor{deepred}{\textbf{67.25}}} & \multicolumn{1}{c}{\textcolor{deepred}{\textbf{70.16}}} & \multicolumn{1}{c}{\textcolor{deepred}{\textbf{-2.62}}}\\

        \bottomrule[1.2pt]
    \end{tabular}}

    \label{tab:internvl_dcl}
\end{table*}

\subsection{Comparison Baselines} \label{exp:baselines}

We compare our method against a set of representative baselines: 

\begin{itemize}
\item \textbf{LoRA-FT} \cite{hu2022lora}: Sequentially updates knowledge through shared low-rank matrices while keeping the pretrained MLLM parameters frozen.

\item \textbf{OLoRA} \cite{wang2023orthogonal}: Mitigates CF by assigning each task to an orthogonal subspace, reducing interference across tasks.

\item \textbf{MoELoRA} \cite{chen2024coin}: Employs multiple independent LoRAs to capture task-specific knowledge from sequential training.

\item \textbf{CL-MoE} \cite{huai2025cl}: Adopts a dual-momentum MoE framework that dynamically selects and updates global and local experts through task-level and instance-level routers, enabling MCIT without CF.

\item \textbf{SEFE} \cite{chensefe}: Tackles two types of forgetting---superficial and essential---by harmonizing task styles via answer style diversification and applying RegLoRA regularization to stabilize key parameters.

\item \textbf{HiDE} \cite{DBLP:conf/acl/GuoZXZWZL25}: Designs a task-specific LoRA expansion and task-general LoRA fusion strategy leveraging layer-wise similarity analysis to balance performance and efficiency while maintaining low memory usage.

\item \textbf{DISCO} \cite{guo2025federated}: Introduces a dynamic knowledge organization mechanism that allocates task-specific LoRA subspaces, with sharing among related tasks, during training, and employs subspace-selective activation at inference time to mitigate cross-task interference.

\end{itemize}

In addition, we consider two reference settings for comparison: \textbf{Zero-shot}: Evaluates each task directly using the pretrained MLLM without additional finetuning. 
\textbf{Individual}: Finetunes the MLLM with LoRA independently on each downstream task, producing a distinct model per task without shared parameters.

\subsection{Evaluation Metrics} 

Following MCITlib \cite{guo2025mcitlib}, we assess MCIT performance with four complementary evaluation metrics.
\textbf{Mean Finetune Accuracy (MFT)} reports the average accuracy obtained on each task immediately after training, reflecting the model's learning ability without the influence of forgetting. 
\textbf{Mean Final Accuracy (MFN)} averages the accuracies of all tasks after the final training stage, indicating how well knowledge is retained overall.
\textbf{Mean Average Accuracy (MAA)} takes the mean of the averaged accuracies across all intermediate training steps, providing a comprehensive view of performance evolution. \textbf{Backward Transfer (BWT)} measures the accuracy difference between the final and post-training states of each task, quantifying the degree of forgetting.

\subsection{Implementation Details}

As PMA is model-agnostic, we incorporate it into two representative MCIT baselines---HiDE and DISCO---to enhance their effectiveness. 
All baselines are built upon widely used MLLMs, including \textbf{LLaVA-1.5-7B} \cite{liu2023improved} and \textbf{InternVL-Chat-7B} \cite{chen2023internvl}, and are trained with LoRA. 
The vision encoder and LLM are frozen, while only the projector and LoRA modules are updated. 
The fused representation has dimension $d_1$+$d_2$, where the CLIP visual embedding dimension is $d_1$=$768$, and the LLM embedding dimension $d_2$ is $4096$ for the 7B backbone. 
Each RD is implemented as a shallow MLP autoencoder with a single bottleneck layer of size ($d_1$+$d_2$)/$4$, trained for $1$ epoch with a learning rate of $1e$-$4$. 
We set the z-score threshold to $\tau$=$1.4$ for distribution-shift detection. 
All experiments are conducted on $4$ NVIDIA A100 GPUs, each with $80$ GB of memory.

\section{Experimental Results} \label{sec:exp_res}

\subsection{Main Results} \label{sub_sec:main_res}

We evaluate PMA on two representative MCIT benchmarks, UCIT and MLLM-DCL, covering diverse task types, instruction styles, and visual distributions. 
Tables \ref{tab:llava_ucit}, \ref{tab:internvl_ucit}, \ref{tab:llava_dcl}, and \ref{tab:internvl_dcl} report the results on LLaVA-1.5-7B and InternVL-Chat-7B backbones across these benchmarks, where PMA is integrated into two strong MCIT baselines, HiDE and DISCO.

\subsubsection{Results on UCIT}
As shown in Tables \ref{tab:llava_ucit} and \ref{tab:internvl_ucit}, incorporating PMA consistently improves performance across all evaluation metrics. 
On the LLaVA-1.5-7B backbone, HiDE+PMA achieves uniform gains across all four MCIT metrics, increasing MFT from $70.24$ to $72.50$, MFN from $64.90$ to $67.87$, and MAA from $78.61$ to $80.57$, while simultaneously mitigating forgetting (BWT: $-5.34 \rightarrow -4.63$). 
When combined with DISCO, PMA further enhances MFT ($75.20 \rightarrow 77.57$), MFN ($68.44 \rightarrow 73.21$), and MAA ($81.36 \rightarrow 83.98$), with a substantially less negative backward transfer (BWT: $-6.76 \rightarrow -4.36$), yielding the best overall performance among all compared methods.

Similar trends are observed on InternVL-Chat-7B. 
HiDE+PMA improves MFT/MFN/MAA/BWT from $73.07$/$68.20$/$81.74$/$-4.88$ to $76.91$/$72.55$/$84.28$/$-4.36$, and DISCO+PMA achieves the SOTA overall results, reaching MFT/MFN/MAA/BWT of $79.04$/$77.42$/$86.59$/$-1.61$.

\subsubsection{Results on MLLM-DCL}
As shown in Tables \ref{tab:llava_dcl} and \ref{tab:internvl_dcl}, PMA consistently improves MCIT performance on the MLLM-DCL benchmark. 
On LLaVA-1.5-7B, HiDE+PMA improves MFT/MFN/MAA from $61.77/56.04/62.30$ to $62.67/57.38/63.49$, while reducing forgetting (BWT: $-5.73 \rightarrow -5.29$). 
When combined with DISCO, PMA further boosts MFT ($64.61 \rightarrow 66.28$), MFN ($59.24 \rightarrow 62.57$), and MAA ($64.01 \rightarrow 65.38$), with a substantially less negative backward transfer (BWT: $-5.37 \rightarrow -3.71$), achieving the best overall performance among all compared methods.

Similar trends are observed on InternVL-Chat-7B. HiDE+PMA improves MFT/MFN/MAA/BWT from $66.47/62.55/67.24/-3.92$ to $68.60/65.05/69.34/-3.55$, while DISCO+PMA attains SOTA overall performance, reaching MFT/MFN/MAA of $69.87/67.25/70.16$ with the lowest forgetting (BWT: $-2.62$).

\subsubsection{Analysis}
Across both benchmarks and backbones, PMA consistently improves MFN and MAA while mitigating forgetting, without sacrificing MFT. These results demonstrate that explicitly addressing projector-level forgetting leads to more stable cross-modal alignment under continual instruction tuning. 
Importantly, the observed gains are orthogonal to backbone-level continual learning strategies such as HiDE and DISCO, indicating that PMA effectively complements existing MCIT methods by targeting an overlooked yet critical source of performance degradation.

\begin{table}[t]
\centering
\caption{Ablation study on UCIT with LLaVA-1.5-7B under the DISCO+PMA setting. The \textbf{bold} denotes the highest result.}
    \resizebox{0.95\linewidth}{!}{
\begin{tabular}{lcccc}
\toprule
Method / Variant & MFT$\uparrow$ & MFN$\uparrow$ & MAA$\uparrow$ & BWT$\uparrow$ \\
\midrule
DISCO* (shared projector) 
& 75.20 & 68.44 & 81.36 & -6.76 \\
\midrule
\textbf{DISCO+PMA (full)} 
& \textbf{77.57} & \textbf{73.21} & \textbf{83.98} & \textbf{-4.36} \\
\midrule
\quad w/o Anchor ($\mathcal{P}^0$ removed) 
& 77.46 & 71.58 & 82.24 & -5.88 \\
\quad No Expansion (single expert) 
& 75.34 & 69.05 & 81.92 & -6.29 \\
\quad Always Expand (one expert per task) 
& 77.53 & 72.37 & 82.84 & -5.16 \\
\quad Top-1 Routing (hard routing) 
& 77.41 & 72.45 & 82.91 & -4.96 \\
\quad Avg Weighting (uniform mixture) 
& 77.18 & 71.91 & 82.43 & -5.27 \\
\bottomrule
\end{tabular}
}
\label{tab:ablation_disco_pma_ucit}
\end{table}

\subsection{Ablation Study} 
We conduct an ablation study based on the DISCO+PMA setting on the UCIT benchmark with the LLaVA-1.5-7B backbone to analyze the contribution of each component. As shown in Table \ref{tab:ablation_disco_pma_ucit}, the full PMA configuration achieves the best performance across all four MCIT metrics. 
Removing the pretrained projector anchor ($\mathcal{P}^0$) preserves strong plasticity but results in noticeably worse MFN and backward transfer, indicating accumulated projector drift without a stable alignment reference. 
Disabling expert expansion causes performance to regress toward the original DISCO baseline, confirming that additional projector capacity is necessary to accommodate multimodal distribution shifts. 
Always expanding a new expert for each task slightly improves MFT but leads to degraded MFN and BWT, suggesting that uncontrolled expansion weakens cross-task sharing. 
Replacing the learned soft router with hard Top-1 routing or uniform averaging also yields inferior MFN and BWT, highlighting the importance of instance-wise, soft expert weighting. 
Overall, while all ablated variants outperform the original DISCO baseline by partially mitigating projector-level forgetting, only the full PMA design consistently achieves a strong balance between plasticity and stability.

\begin{figure}[t!]
\centering
  \includegraphics[width=1.0\linewidth]{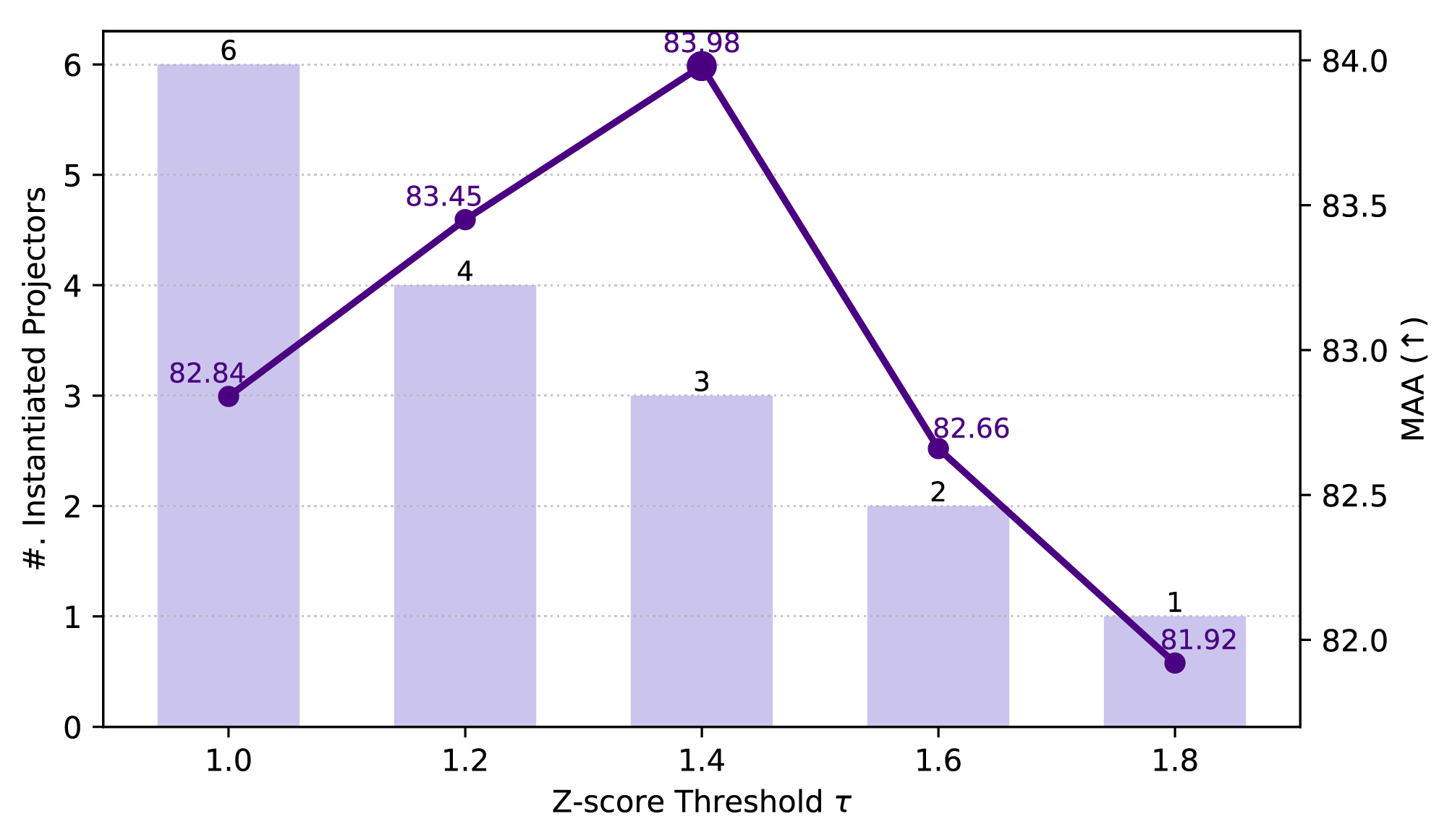}
  \caption{Effect of the z-score threshold $\tau$ on performance and projector growth under the DISCO+PMA setting on UCIT with LLaVA-1.5-7B. An intermediate threshold ($\tau=1.4$) achieves the best MAA performance with sub-linear growth.}
\label{fig:tau_ucit}
\end{figure}

\subsection{Hyperparameter Analysis}

We study the effect of the distribution-shift (z-score) threshold $\tau$ in PMA under the DISCO+PMA setting on UCIT with LLaVA-1.5-7B, focusing on the trade-off between continual performance and parameter growth. 
Figure \ref{fig:tau_ucit} reports the resulting MAA and the number of instantiated projectors across different $\tau$ values. 
A smaller threshold (\emph{e.g.}, $\tau$=$1.0$) makes the detector overly sensitive, triggering expansions at nearly every task and leading to rapid parameter growth ($6$ experts), which weakens expert reuse and cross-task knowledge sharing, resulting in suboptimal performance (MAA=$82.84$). 
In contrast, a large threshold (\emph{e.g.}, $\tau$=$1.8$) rarely triggers expansion, yielding minimal growth ($1$ expert) but insufficient adaptation, degrading performance (MAA=$81.92$). 
Intermediate thresholds strike a better balance between efficiency and adaptability. 
In particular, $\tau$=$1.4$ achieves the highest MAA ($83.98$) with only $3$ experts, demonstrating that PMA attains strong MCIT performance with sub-linear projector growth by expanding capacity only when necessary.

\begin{figure}[t!]
\centering
  \includegraphics[width=1.0\linewidth]{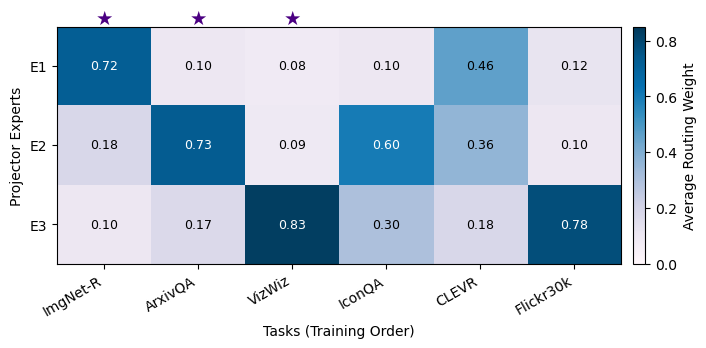}
\caption{Projector usage analysis under the DISCO+PMA setting on UCIT with LLaVA-1.5-7B. 
Early tasks trigger projector expert expansion and mainly use their newly created experts, while later tasks reuse previously learned experts.}
\label{fig:projector_usage_ucit}
\end{figure}

\subsection{Projector Usage Analysis} 

We analyze the projector expansion and reuse behavior of PMA under DISCO+PMA on UCIT with LLaVA-1.5-7B. 
After completing training on all $6$ UCIT tasks, we evaluate the final model and visualize how PMA routes samples from each task's test set to different projector experts. 
Figure \ref{fig:projector_usage_ucit} reports the average routing weights of each task over the learned experts at the final evaluation stage. 
With $\tau$=$1.4$, PMA instantiates $3$ projector experts during training. 
The first three tasks (ImgNet-R, ArxivQA, and VizWiz), which trigger expert expansion, predominantly rely on the experts created for them, reflecting clear task-specific alignment needs as well as the bias toward newly added projectors described in Section \ref{sub_sec:expansion}. Importantly, later tasks exhibit substantial expert reuse. 
IconQA mainly reuses the expert trained on ArxivQA, while Flickr30k primarily reuses the expert trained on VizWiz, indicating that PMA captures transferable cross-modal alignment patterns among tasks with similar instruction styles and visual distributions. 
Notably, CLEVR shows mixed usage across multiple experts, which is consistent with PMA's soft reuse mechanism in Section \ref{sub_sec:expansion}, where the router assigns higher weights to the most compatible prior experts while allowing non-zero contributions from others. 
Since CLEVR combines short-answer outputs with structured reasoning, it benefits from integrating experts learned from both natural-image and structured QA tasks. 
Overall, these results show that PMA expands projector capacity only when necessary and achieves efficient sub-linear growth by flexibly reusing previously learned projectors.

\section{Conclusion}
In this paper, we identify projector-level forgetting as a critical yet largely overlooked challenge in MCIT. 
While existing approaches mainly focus on mitigating CF within the LLM backbone, we show that the shared projector responsible for cross-modal alignment can drift under sequential updates, leading to degraded instruction-following performance on previously learned tasks. 
To address this issue, we propose PMA, a method-agnostic framework that enables continual adaptation of the projector while preserving previously learned alignments. PMA detects multimodal distribution shifts using lightweight RDs, expands projector experts only when necessary, and integrates them via an expandable router anchored by the original pretrained projector. 
Extensive experiments on two MCIT benchmarks demonstrate that explicitly modeling projector-level adaptation consistently improves SOTA methods and scales effectively across different MLLM backbones.

\section*{Acknowledgment}
This work was supported by the Brain Science and Brain-like Intelligence Technology - National Science and Technology Major Project (2025ZD0217200), Strategic Priority Research Program of the Chinese Academy of Sciences (Grant No. XDB1010302), CAS Project for Young Scientists in Basic Research (YSBR-116), Youth Innovation Promotion Association CAS, Shanghai Leading Talent Program of Eastern Talent Plan, the Lingang Laboratory Fund (Grant No. LG-GG-202402-06-07, LGL-1987-09), the Shanghai Municipal Science and Technology Project (Grant No. 25ZR1401370, 25LN3200400), Special Support Project of Guangdong Province (Grant No.0720240209). 
The numerical calculations in this study were carried out on the ORISE Supercomputer.

\bibliographystyle{ACM-Reference-Format}
\bibliography{sample-base}

\end{document}